\documentclass[11pt]{article}

\usepackage[final]{acl}

\usepackage{times}
\usepackage{latexsym}

\usepackage[T1]{fontenc}
\usepackage[utf8]{inputenc}

\usepackage{microtype}

\usepackage{inconsolata}

\usepackage{graphicx}
\usepackage{amsmath}
\usepackage{enumitem}
\usepackage{booktabs}
\usepackage{multirow}
\usepackage{colortbl}
\usepackage{amssymb}
\usepackage{tabularx}
\usepackage{array}

\title{Rethinking Table Pruning in TableQA: From Sequential Revisions to Gold Trajectory-Supervised Parallel Search}

\author{
  \textbf{Yu Guo\textnormal{\textsuperscript{1}\footnotemark[1]}},
  \textbf{Shenghao Ye\textnormal{\textsuperscript{1}\footnotemark[1]}},
  \textbf{Shuangwu Chen\textnormal{\textsuperscript{1}\footnotemark[2]}},
  \textbf{Zijian Wen\textnormal{\textsuperscript{1}}},
  \textbf{Tao Zhang\textnormal{\textsuperscript{1}}},
  \textbf{Qirui Bai\textnormal{\textsuperscript{1}}}\\
  \textbf{Dong Jin\textnormal{\textsuperscript{2}\footnotemark[2]}},
  \textbf{Yunpeng Hou\textnormal{\textsuperscript{2}}},
  \textbf{Huasen He\textnormal{\textsuperscript{1}}},
  \textbf{Jian Yang\textnormal{\textsuperscript{1}}},
  \textbf{Xiaobin Tan\textnormal{\textsuperscript{1}}}
\\
  \textsuperscript{1}University of Science and Technology of China
\\
  \textsuperscript{2}Institute of Artificial Intelligence, Hefei Comprehensive National Science Center
\\
  \texttt{\{yukariguo, ssh0321y, wzj20020304, zhangtaolqy, bqr135, hyp314\}@mail.ustc.edu.cn}
\\
  \texttt{\{chensw, kingdon, hehuasen, jianyang, xbtan\}@ustc.edu.cn}
}

\begin{document}
\maketitle

\renewcommand{\thefootnote}{\fnsymbol{footnote}}
\footnotetext[1]{Equal contribution}
\footnotetext[2]{Corresponding authors}
\begin{abstract}

Table Question Answering (TableQA) benefits significantly from table pruning, which extracts compact sub-tables by eliminating redundant cells to streamline downstream reasoning. However, existing pruning methods typically rely on sequential revisions driven by unreliable critique signals, often failing to detect the loss of answer-critical data. To address this limitation, we propose TabTrim, a novel table pruning framework which transforms table pruning from sequential revisions to gold trajectory-supervised parallel search. TabTrim derives a gold pruning trajectory using the intermediate sub-tables in the execution process of gold SQL queries, and trains a pruner and a verifier to make the step-wise pruning result align with the gold pruning trajectory. During inference, TabTrim performs parallel search to explore multiple candidate pruning trajectories and identify the optimal sub-table. Extensive experiments demonstrate that TabTrim achieves state-of-the-art performance across diverse tabular reasoning tasks: TabTrim-8B reaches 73.5\% average accuracy, outperforming the strongest baseline by 3.2\%, including 79.4\% on WikiTQ and 61.2\% on TableBench.

%TabTrim leverages text-to-SQL datasets to derive step-level gold sub-table trajectories by decomposing gold SQL and executing clause-level operations, and constructs off-trajectory negatives without additional annotation. We then train (1) a Trajectory-supervised Pruner with trajectories supervision and preference optimization, and (2) a Loss-aware Verifier that regresses a recall-biased quality score to explicitly penalize evidence loss. At inference, TabTrim performs Parallel Trajectory Search with a beam-style propose–score–select procedure to maintain multiple competing pruning paths and output the highest-scoring sub-table for downstream reasoning. Extensive experiments show that TabTrim achieves consistent gains across diverse tabular reasoning tasks: TabTrim-8B reaches 73.5\% average accuracy, outperforming the strongest baseline by 3.2\%, including 79.4\% on WikiTQ and 61.2\% on TableBench.

\end{abstract}

\section{Introduction}

\begin{figure}[!t]
  \includegraphics[width=\linewidth]{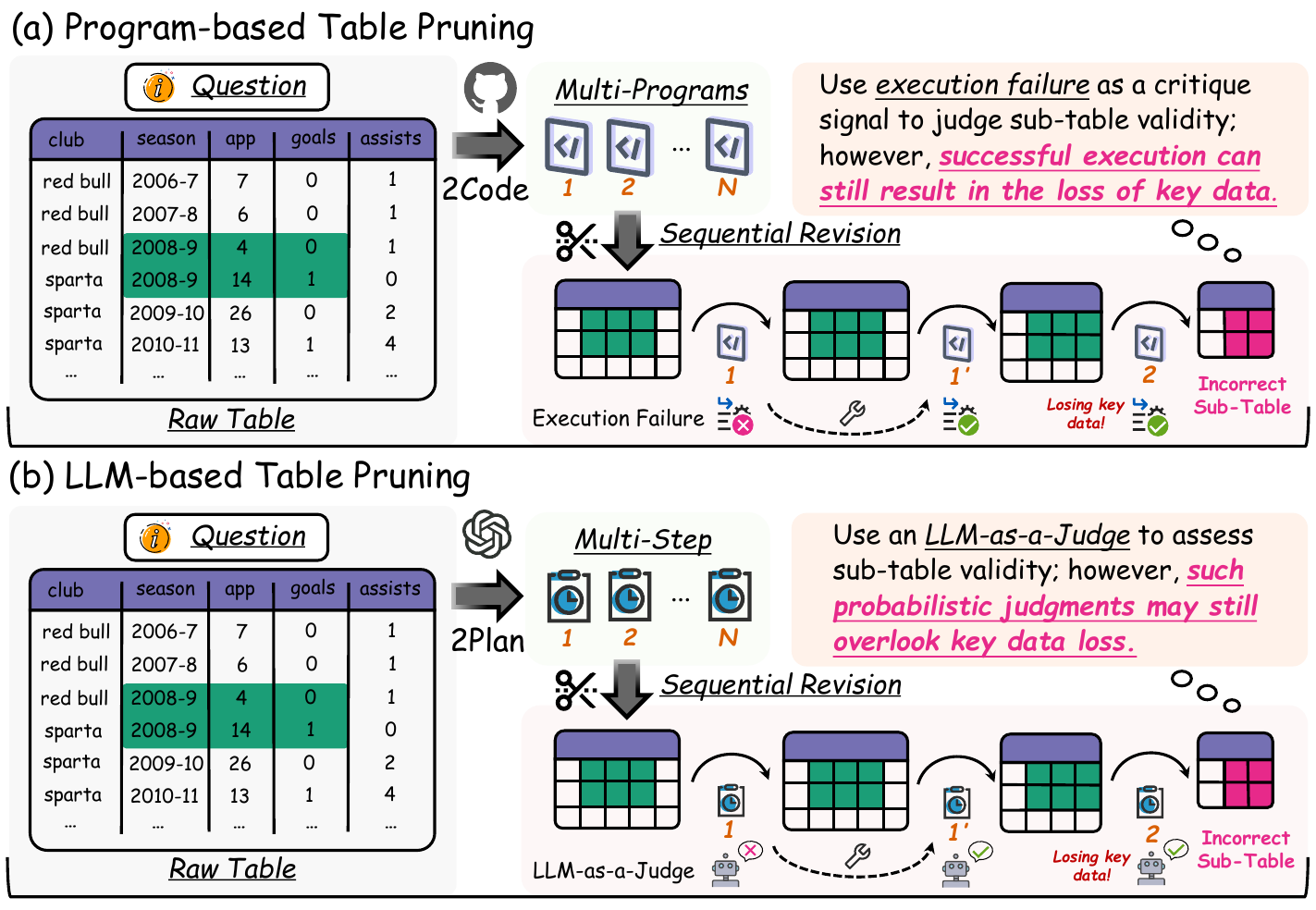}
  \caption{Illustration of (a) Program-based and (b) LLM-based pruning paradigms. \includegraphics[width=0.3cm]{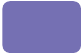} represents table headers, \includegraphics[width=0.3cm]{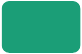} represents  gold cells related to answer and \includegraphics[width=0.3cm]{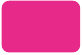} represents  incomplete gold cells.}
  \label{intro}
\end{figure}

\begin{figure*}[!t]
  \includegraphics[width=\linewidth]{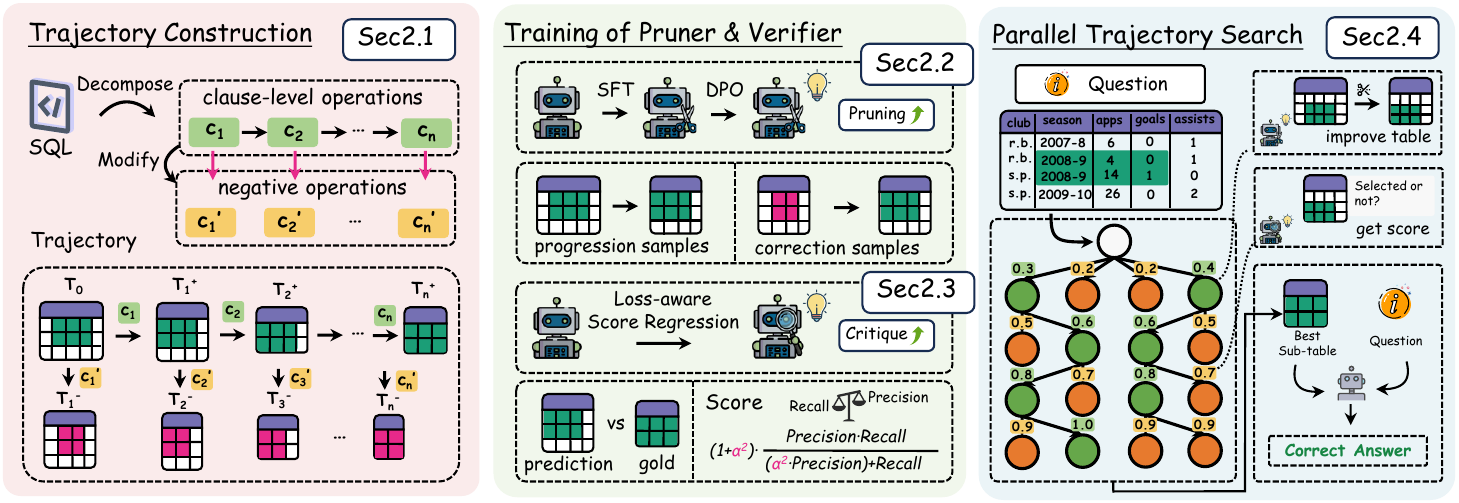}
  \caption{Overview of TabTrim. TabTrim constructs gold sub-table trajectories to provide supervision for two core components: a Trajectory-supervised Pruner and a Loss-aware Verifier. During inference, TabTrim performs Parallel Trajectory Search with a step-wise generate–score–select procedure: the pruner generates candidate pruned sub-tables, the verifier assesses sub-tables with loss-aware scores, and the search retains the top candidates, finally outputting the highest-scoring sub-table for downstream reasoning.}
  \label{method}
\end{figure*}

% So how can we provide reliable and grounded supervision to prevent answer-critical data loss during multi-step table pruning?

Table Question Answering (TableQA), which enables users to retrieve information from tabular data via natural language \citep{zhang2025survey}, remains a non-trivial challenge, as numerous irrelevant rows and columns in raw tables often disrupt reasoning process \citep{chen2024tablerag}. Against this backdrop, table pruning \citep{wu-etal-2023-tacr, lin-etal-2023-inner}, which seeks to extract compact sub-tables by removing redundant cells, has emerged as a promising approach to streamline downstream  TableQA.
%by extracting compact sub-tables and removing redundant table data

Existing pruning methods generally fall into two paradigms, as illustrated in Fig.\ref{intro}. 
Program-based methods \citep{Binder, 10.14778/3659437.3659452, nahid-rafiei-2024-tabsqlify} parse input query and invoke executable programs (e.g., SQL and Python) to convert pruning tasks into  procedural operations.
% 根据问题生成multi program，pruning Table step-by-step
In contrast, LLM-based methods \citep{ye2023large, wangchain} use Chain-of-Thought or multi-agent planning to perform pruning through multi-step reasoning.
Despite promising results, both paradigms are prone to errors in pruning, particularly for intricate queries or tables where the implicit associations between query semantics and tabular data become obscure. 
%Accordingly, an early pruning error is likely to cause the loss of answer-critical data, which significantly degrades TableQA performance  \citep{yu-etal-2025-table}.
These errors  may propagate and accumulate in multi-step pruning,  significantly degrading TableQA performance  \citep{yu-etal-2025-table}.

To tackle this issue, recent studies  introduce critique signals to refine  pruning decisions at each step for both paradigms.
%However, they still fail to reliably prevent the loss of answer-critical data. 
For program-based pruning, execution failures are used to trigger program repair or regeneration \citep{jin-etal-2025-talon}. 
They can capture  syntactic or runtime errors yet scarcely pinpoint  semantic errors, as a program may execute successfully but  remove answer-critical data.
For LLM-based pruning, an LLM-as-a-Judge is often employed to check whether the pruned cells are  redundant at each reasoning step \citep{yu-etal-2025-table}.
However,  mandating LLMs to perform  self-reflection readily induces subjective bias, as models tend to  either rationalize their  erroneous reasoning or over-criticize their   valid steps  \citep{huanglarge}, leading to inconsistent judgments.
%%leading to inconsistent judgments of sub-table utility
%leading to over- or under-pruning of  answer-critical  data.
 
Rethinking table pruning in TableQA, we attribute these limitations to two fundamental  reasons:
\textbf{(1) Unreliable Critique.}  Without an explicit objective grounding, existing critique mechanisms can barely verify whether we have over-pruned critical cells or under-pruned redundant content at each  step, thus undermining the reliability of table pruning.
\textbf{(2) Sequential Revision.} 
Existing multi-step table pruning  is inherently a sequential revision process that iteratively refines pruning results based on the reflection of preceding iterations.
Such refinement is usually confined to a single pruning trajectory, unable to backtrack and explore alternatives, thereby readily getting trapped in suboptimal pruning results.

To address these challenges,  we identify  two key insights:
(1) % Although existing TableQA datasets  lack step-wise sub-table annotations, 
The intermediate sub-tables yielded from  gold SQL execution for Text-to-SQL tasks 
%form a gold pruning trajectory, and thus can  serve as a reliable objective critique for  each pruning step.
are grounded on verified query results, and thus can  serve as a reliable gold pruning trajectory for critiquing each  pruning step.
(2)  By generating multiple candidate pruning trajectories in parallel and searching for the optimal one, the exploration space can be effectively expanded, thereby avoiding local optima.

In light of the above insights, we propose \textbf{TabTrim}, a novel  gold trajectory-supervised parallel search framework for table pruning.
%searching optimal pruned sub-tables by aligning critique signals and enabling multi-path pruning.
Specifically, TabTrim derives a gold  pruning trajectory using the intermediate sub-tables in the execution process of  gold SQL queries, and trains a pruner and a verifier to make the  step-wise pruning  result align with the  gold  pruning trajectory.
During inference, TabTrim performs parallel search  to explore multiple candidate pruning trajectories and identify the optimal  sub-table.

\textbf{Our Contributions.}
(1) \textit{New Insight.} We rethink table pruning in TableQA by transforming it from sequential revisions to gold trajectory-supervised parallel search.
(2) \textit{Novel Framework.} We propose TabTrim, a novel table pruning framework which leverages  gold trajectory as a reliable critique for  each pruning step and employs parallel search to find the optimal sub-table.
(3) \textit{SOTA Performance.} Extensive experiments demonstrate that TabTrim achieves state-of-the-art performance across diverse tabular reasoning tasks. TabTrim-8B reaches 73.5\% average accuracy, outperforming the strongest baseline by 3.2\%, including 79.4\% on WikiTQ and 61.2\% on TableBench.

\section{TabTrim}

% In this section, we introduce TabTrim, a novel framework for multi-step sub-table generation under complex TableQA settings, as depicted in Fig.~\ref{method}. TabTrim is designed to prevent the loss of answer-related data during sub-table generation by explicitly assessing sub-table quality and performing iterative self-correction.

In this section, we introduce TabTrim, a framework that performs step-wise pruning with reliable critique through parallel search, reducing answer-critical data loss while extracting simplified sub-tables to enhance downstream reasoning. We begin by introducing the trajectories construction (Sec.\ref{sec:data}), which serves as the supervision for the two core components of TabTrim, as shown in Fig.\ref{method}: (1) Trajectories-supervised Pruner, which generates pruned sub-tables step by step (Sec.\ref{sec:pruner}); and (2) Loss-aware Verifier, which assesses each sub-table with a loss-aware score (Sec.\ref{sec:verifier}). Finally, TabTrim performs inference via Parallel Trajectory Search, which  invokes the pruner to generate multiple candidate sub-tables, rejects low-quality ones using verifier scores, and outputs the highest-scoring sub-table for downstream  table reasoning (Sec.\ref{sec:search}).

\subsection{Trajectory Construction}
\label{sec:data}

Since existing TableQA datasets only provide final answers and lack step-level sub-table annotations, we derive sub-table trajectories from single-table Text-to-SQL data by executing a sequence of decomposed gold SQL clause-level operations. The constructions of gold sub-table trajectories and off-trajectory negatives are detailed as follows.
%Specifically, we construct gold sub-table trajectories and off-trajectory negatives.

\paragraph{Gold Trajectories via SQL Decomposition.} 

We leverage existing Text-to-SQL datasets $\mathcal{D}=\{(Q, SQL_{\text{gold}}, T_{\text{raw}})\}$ as data sources, where each instance consists of a natural language question $Q$, an annotated gold SQL query $SQL_{\text{gold}}$, and the corresponding raw table $T_{\text{raw}}$.  Exploiting the compositional structure of SQL, we decompose the gold query $SQL_{\text{gold}}$ into a sequence of clause-level operations $(c_1, c_2, \dots, c_n)$ following the query’s logical execution order. Here, we focus on clause-level operations that directly affect answer-critical data, such as row filtering and column projection (where projection retains columns referenced by \texttt{SELECT}, \texttt{GROUP BY}, \texttt{HAVING}, or aggregate expressions). By sequentially executing these operations, we obtain a gold sub-table trajectory $\tau^{+}=(T_0, T_1^{+}, \dots, T_n^{+})$, where $T_t^{+}$
denotes the gold sub-table after executing $c_t$ at step $t=1,\dots,n$ and $T_0$ is initialized by $T_{\text{raw}}$.
This procedure produces gold sub-table trajectories without requiring additional manual annotation. More details about SQL decomposition are shown in Appendix~\ref{sec:decomposition}.

\paragraph{Off-trajectory Negatives Construction.}
To improve model's robustness against erroneous pruning, we construct negative sub-tables off the gold trajectory by modifying gold  clause-level operations.
%To cover off-trajectory sub-table negatives that may arise at inference and equip the pruner with recovery capability, we construct negative samples by modifying gold clause components. 
Executing the modified operations on $T_{t-1}^+$ yields non-trivial but incorrect sub-tables $T_t^{-}$. 
The training samples are formed as progression dataset $\mathcal{D}^+=\{(Q, T_0, T_{t-1}^+, T_t^{+})\}$ and correction dataset $\mathcal{D}^-=\{(Q, T_0, T_{t-1}^-, T_t^{+})\}$. More details about SQL modification are shown in Appendix~\ref{sec:modification}.

\subsection{Gold Trajectory-supervised Pruner}
\label{sec:pruner}

We use the trajectories constructed above as  supervision signals to train the pruner $M_{\theta}$, which generates the next pruned sub-table $T_t$ conditioned on the question $Q$, the raw table $T_{\text{raw}}$, and the current sub-table $T_{t-1}$. The training process is conducted in two distinct stages.

\paragraph{Stage 1: Supervised Fine-Tuning.}

To teach the pruner to align with the gold pruning trajectory and equip it with the ability to recover from erroneous pruning, we include both progression samples from $\mathcal{D}^+$ and correction samples from $\mathcal{D}^-$. We then optimize the pruner with the following supervised fine-tuning objective:
\begin{equation}
    \scalebox{0.78}{$
    \begin{split}
    \mathcal{L}_\text{SFT}(\theta)
    = - \sum_{(Q,T_0,T_{t-1}^+,T_t^+) \in \mathcal{D}^+} \log P_\theta(T_t^+ \mid Q, T_0, T_{t-1}^+) \\
    \qquad\;\; - \lambda \sum_{(Q,T_0,T_{t-1}^-,T_t^+) \in \mathcal{D}^-} \log P_\theta(T_t^+ \mid Q, T_0, T_{t-1}^-),
    \end{split}
    $}
\end{equation}
where $\theta$ denotes the model parameters and $\lambda$ weights the correction term. The first term of loss function trains the pruner to follow the gold trajectory from $T_{t-1}^{+}$ to $T_t^{+}$, while the second term trains it to recover to the same target $T_t^{+}$ from a modified sub-table $T_{t-1}^{-}$. The sums range over all training instances and time steps $t$.

\paragraph{Stage 2: Direct Preference Optimization.}
Supervised fine-tuning alone may still yield sub-tables with subtle semantic errors. We therefore employ Direct Preference Optimization (DPO) \citep{rafailov2023direct} to enforce a preference for the gold next sub-table $T_t^+$ over the incorrect next sub-table $T_t^-$ under the same context. The loss is:
\begin{equation}
    \scalebox{0.85}{$
    \begin{split}
    \mathcal{L}_\text{DPO}(\theta)
    = - \log \sigma \Big(
    \beta \log \frac{P_\theta(T_t^+ \mid Q, T_0, T_{t-1})}{P_\text{ref}(T_t^+ \mid Q, T_0, T_{t-1})} \\
    \qquad\qquad\qquad\quad
    - \beta \log \frac{P_\theta(T_t^- \mid Q, T_0, T_{t-1})}{P_\text{ref}(T_t^- \mid Q, T_0, T_{t-1})}
    \Big),
    \end{split}
    $}
\end{equation}
where $P_\theta$ denotes the target model to be optimized and $P_{\text{ref}}$ is a frozen reference model. The two log-ratio terms compare how much the target model increases the likelihood of $T_t^{+}$ and decreases the likelihood of $T_t^{-}$ relative to the reference model under the same context. $\beta$ controls the strength of this preference, and $\sigma(\cdot)$ is the logistic sigmoid. This stage pushes the model to prefer the gold pruning process $T_{t-1}\!\to\!T_t^{+}$ over the negative pruning process $T_{t-1}\!\to\!T_t^{-}$, reducing subtle errors.

\subsection{Loss-aware Verifier}
\label{sec:verifier}

To quantify how far a pruned sub-table deviates from the optimal one, we introduce a Loss-aware Verifier $g_\phi$ that assesses each sub-table with a quality score.

\paragraph{Loss-aware Quality Score.}

To obtain an objective supervision, we compare each intermediate sub-table $T_t$ with the optimal reference sub-table, defined as the final sub-table $T_n^+$ of the gold trajectory constructed in Sec.~\ref{sec:data}.

To make this comparison computable, we canonicalize the raw table $T_{\text{raw}}$ into a set of indexed cells and represent its sub-table as a subset of  canonical cell set. Concretely, each cell is identified by its row index $r$, column index $c$, and value $v_{rc}$. 
Each sub-table is represented by a set of selected cells $E(T)=\{(r,c,v_{rc})\}$. More details about canonicalization are shown in \ref{sec:canonicalization}.

We then define Precision and Recall by cell-set overlap between the intermediate sub-table and the optimal reference sub-table:
\begin{equation}
    \begin{split}
    \text{Precision}=\frac{|E(T_t)\cap E(T_n^+)|}{|E(T_t)|}, \\
    \text{Recall}=\frac{|E(T_t)\cap E(T_n^+)|}{|E(T_n^+)|}.
    \end{split}
\end{equation}
Intuitively, Recall measures how much answer-critical data is preserved, while Precision reflects the amount of redundant cells retained. Since losing answer-critical data is typically more detrimental to downstream reasoning, we define the loss-aware quality score as an $\alpha$-weighted F-score:
\begin{equation}
    \scalebox{0.85}{$
    \begin{split}
    S(T_t)
    = (1 + \alpha^2) \cdot \frac{\text{Precision} \cdot \text{Recall}}
    {(\alpha^2 \cdot \text{Precision}) + \text{Recall}}.
    \end{split}
    $}
\end{equation}
Here, $\alpha$ controls the penalty on answer-critical data loss. In particular, choosing $\alpha>1$ emphasizes recall, so missing answer-critical data leads to a larger score drop.

\paragraph{Training.}
We train the verifier to regress from $(Q, T_0, T_t)$ to the loss-aware quality score $S(T_t)$. Specifically, we construct a verifier training set $\mathcal{D}_{\text{ver}}$ using sub-tables from both (1) gold trajectories $\{T_t^+\}$ and (2) modified off-trajectory negatives $\{T_t^-\}$ (Sec.~\ref{sec:data}). For each sub-table $T_t$, we compute its score by comparing it with $T_n^+$, and then optimize the verifier with the following objective:
\begin{equation}
    \scalebox{0.85}{$
    \begin{split}
    \mathcal{L}_{\text{ver}}(\phi)
    = \sum_{(Q, T_0, T_t)\in \mathcal{D}_{\text{ver}}}
    \big(g_\phi(Q, T_0, T_t) - S(T_t)\big)^2.
    \end{split}
    $}
\end{equation}
At inference time, the learned verifier predicts step-level loss-aware scores for sub-tables without access to $T_n^+$.

\subsection{Parallel Trajectory Search}
\label{sec:search}

TabTrim performs inference via parallel search, which incrementally constructs pruning trajectories as sequences of sub-tables. Starting from the initial table $T_0$, at each step the pruner $M_\theta$ generates multiple candidate next sub-tables for each currently retained sub-table. The verifier $g_\phi$ then assesses a step-level loss-aware score to each candidate, and TabTrim selects a small set of top-ranked candidates to continue expanding in the next step. Unlike sequential revisions that commit to single pruning trajectory, TabTrim maintains multiple competing sub-table trajectories and performs a generate--score--select procedure step by step. This design reduces the risk of early pruning errors by allowing the search to discard low-quality branches while preserving promising alternatives.

Specifically, we implement parallel search process via beam search. Let $k$ denote the beam width, $b$ the branching factor, and $D_{\max}$ the maximum depth.
We initialize $\mathcal{B}_0=\{T_0\}$.
For $t=0$, we warm-start by sampling $k \cdot b$ candidates from $T_0$ and selecting the top-$k$ sub-tables; for $t\ge 1$, each beam sub-table proposes $b$ candidates.
For each step $t=0,\ldots,D_{\max}-1$, we form the candidate pool
\begin{equation}
\scalebox{0.75}{$
\mathcal{U}_{t+1}=\{\tilde{T}_{t+1}^{j,i}=M_\theta(Q,T_0,\tilde{T}_t^{j}) \mid \tilde{T}_t^{j}\in \mathcal{B}_t,\ i=1,\ldots,b\},
$}
\end{equation}
score each candidate with the verifier
\begin{equation}
\hat{S}_\phi(\tilde{T}_{t+1}^{j,i}) = g_\phi(Q,T_0,\tilde{T}_{t+1}^{j,i}),
\end{equation}
and keep the top-$k$ candidates as the next beam:
\begin{equation}
\mathcal{B}_{t+1}=\mathrm{TopK}(\mathcal{U}_{t+1},k;\hat{S}_\phi).
\end{equation}
Finally, we output the highest-scoring sub-table among all explored beams:
\begin{equation}
\hat{T}=\arg\max_{\tilde{T}\in \bigcup_{t=0}^{D_{\max}}\mathcal{B}_t} \hat{S}_\phi(\tilde{T}),
\end{equation}
and feed $(Q,\hat{T})$ to an LLM to generate the answer $A$.

\begin{table*}[t]
  \centering
  \small
  \setlength{\tabcolsep}{4pt}
  \resizebox{\textwidth}{!}{
  \begin{tabular}{lcccccc}
    \toprule
    \textbf{Method} &
    \textbf{WikiTQ} & \textbf{TabFact} &
    \textbf{TB-NR} & \textbf{TB-FC} & \textbf{TB-DA} &
    \textbf{Average} \\
    \midrule
    \multicolumn{7}{c}{\textit{\textbf{Direct QA}}}\\
    Qwen3-4B~\citep{yang2025qwen3}          & 50.8 & 74.1 & 61.7 & 70.8 & 19.2 & 55.3 \\
    Qwen3-8B~\citep{yang2025qwen3}          & 52.2 & 76.7 & 64.5 & 72.9 & 23.0 & 57.9 \\
    GPT-4o-mini~\citep{hurst2024gpt}        & 54.3 & 77.4 & 65.5 & 76.0 & 25.1 & 59.8 \\
    \midrule
    \multicolumn{7}{c}{\textit{\textbf{Program-based methods}}}\\
    Binder~\citep{Binder}                  & 54.8 & 83.3 & 66.8 & 67.7 & 26.8 & 59.9 \\
    TabSQLify~\citep{nahid-rafiei-2024-tabsqlify} & 68.7 & 78.3 & 65.2 & 76.0 & 28.0 & 63.2 \\
    \midrule
    \multicolumn{7}{c}{\textit{\textbf{LLM-based methods}}}\\
    Dater~\citep{ye2023large}              & 65.8 & 83.6 & 65.0 & 69.8 & 28.6 & 62.6 \\
    Chain-of-Table~\citep{wangchain}       & 67.5 & 88.9 & 68.5 & 78.1 & 30.3 & 66.7 \\
    \midrule
    \multicolumn{7}{c}{\textit{\textbf{Critique methods}}}\\
    TALON~\citep{jin-etal-2025-talon}      & 70.7 & 87.6 & 67.3 & 77.1 & 28.9 & 66.3 \\
    Table-Critic~\citep{yu-etal-2025-table}& 72.6 & \underline{90.6} & 73.0 & \underline{81.3} & \underline{33.8} & 70.3 \\
    \midrule
    \multicolumn{7}{c}{\textit{\textbf{Ours}}}\\
    \rowcolor[HTML]{D8ECE4}
    TabTrim-4B           & \underline{76.8} & 89.4 & \underline{76.3} & 79.2 & 32.1 & \underline{70.8} \\
    \rowcolor[HTML]{D8ECE4}
      TabTrim-8B           & \textbf{79.4} & \textbf{91.2} & \textbf{78.8} & \textbf{83.3} & \textbf{34.7} & \textbf{73.5} \\
    % \textbf{TabTrim-8B}           & \textbf{79.4\scriptsize{\textcolor[HTML]{076B3D}{(+6.8)}}} & \textbf{91.2\scriptsize{\textcolor[HTML]{076B3D}{(+0.6)}}} & \textbf{78.8\scriptsize{\textcolor[HTML]{076B3D}{(+5.8)}}} & \textbf{83.3\scriptsize{\textcolor[HTML]{076B3D}{(+2.0)}}} & \textbf{34.7\scriptsize{\textcolor[HTML]{076B3D}{(+0.9)}}} & \textbf{73.5\scriptsize{\textcolor[HTML]{076B3D}{(+3.2)}}} \\
    \bottomrule
  \end{tabular}
  }
  \caption{Main performance comparison on WikiTQ, TabFact and TableBench benchmarks.
  \textbf{Bold} denotes the best performance and \underline{underline} denotes the second-best performance.}
  \label{tab:main_results}
\end{table*}

\section{Experiments}

\subsection{Experimental Setup}

\paragraph{Models.}

We construct over 80K training samples from WikiSQL~\citep{zhong2017seq2sql} and SQUALL \citep{shi-etal-2020-potential} using the data construction procedure described in Sec.\ref{sec:data}. Based on this data, we train Qwen3-4B and Qwen3-8B as pruners, and train Qwen3-0.6B as the verifier with $\alpha=1.5$. Unless otherwise stated, we set the beam width and branching factor to $k=b=2$ and the maximum search depth to $D_{\max}=4$, which yields an upper bound of $O(k \cdot b \cdot D_{\max})$ pruner and verifier calls per example. This results in TabTrim-4B and TabTrim-8B. More training and implementation details are provided in Appendix~\ref{sec:setup}.

\paragraph{Datasets.}

We evaluate our method on three representative and challenging benchmarks spanning diverse tabular reasoning tasks, including (1) WikiTQ \citep{pasupat-liang-2015-compositional}, a table reasoning dataset with 4,344 samples from 421 Wikipedia tables. (2) TabFact \citep{2019TabFactA}, a fact verification benchmark in table reasoning with 2,024 test samples from 298 tables. (3) TableBench (TB) \citep{wu2025tablebench}, a complex tabular reasoning benchmark with 886 questions covering tasks of numerical reasoning (NR), fact checking (FC) and data analysis (DA). More information about datasets is shown in Appendix~\ref{sec:dataset}.

\paragraph{Evaluation Metric.}

For WikiTQ and TableBench datasets, we use exact match accuracy (EM) to check whether the predicted answer matches the ground truth. For TabFact, we adopt binary classification accuracy as evaluation metric.

\paragraph{Compared Methods.}

We compare TabTrim with baselines from four categories: \textbf{(1) Direct QA:} Qwen3~\citep{yang2025qwen3} and GPT-4o-mini~\citep{hurst2024gpt}. \textbf{(2) Program-based methods:} Binder~\citep{Binder} and TabSQLify~\citep{nahid-rafiei-2024-tabsqlify}. \textbf{(3) LLM-based methods:} Dater~\citep{ye2023large} and Chain-of-Table~\citep{wangchain}. \textbf{(4) Critique methods:} TALON~\citep{jin-etal-2025-talon} and Table-Critic~\citep{yu-etal-2025-table}. Since many baselines couple pruning and reasoning within closed-source frameworks, we standardize our evaluation by executing their complete pipelines with GPT-4o-mini. For each method, we strictly follow its original settings to ensure peak performance. For TabTrim, we also use GPT-4o-mini to generate the final answer from the selected sub-table to ensure a fair comparison.

\subsection{Main Results}

Tab.\ref{tab:main_results} reports the performance of TabTrim and all baselines on WikiTQ, TabFact and TableBench. Our comprehensive evaluation reveals several key findings. First, TabTrim achieves the best overall performance across all benchmarks. TabTrim-8B reaches an average accuracy of 73.5\%, outperforming the strongest non-TabTrim baseline (Table-Critic, 70.3\%) by 3.2\%, while TabTrim-4B also attains a strong 70.8\% average and already surpasses all non-TabTrim methods. Second, TabTrim delivers consistent gains across diverse tabular reasoning tasks. On WikiTQ, TabTrim-8B achieves 79.4\%, exceeding the best baseline by 6.8\%, highlighting that our step-wise pruning coupled with reliable verification is especially effective for complex compositional questions. On TabFact, TabTrim-8B achieves high performance of 91.2\%, indicating that the advantage of improved pruning extends beyond only hard instances. Third, TabTrim generalizes well to TableBench, where it achieves the best performance on all subtasks. Compared with the strongest baseline, it improves TB-NR by 5.8\%, TB-FC by 2.0\%, and TB-DA by 0.9\%. The especially large gain on TB-NR suggests that TabTrim is particularly effective for multi-hop numerical reasoning, where correct answering is highly sensitive to retaining the right operands and critical data. Finally, we observe a clear scaling effect with stronger pruners. Upgrading the pruner from 4B to 8B consistently improves performance across all datasets, yielding a 2.5\% average gain, which suggests that better proposal quality further amplifies the benefit of Parallel Trajectory Search under the verification.

\begin{figure}[t]
    \centering
    \includegraphics[width=\linewidth]{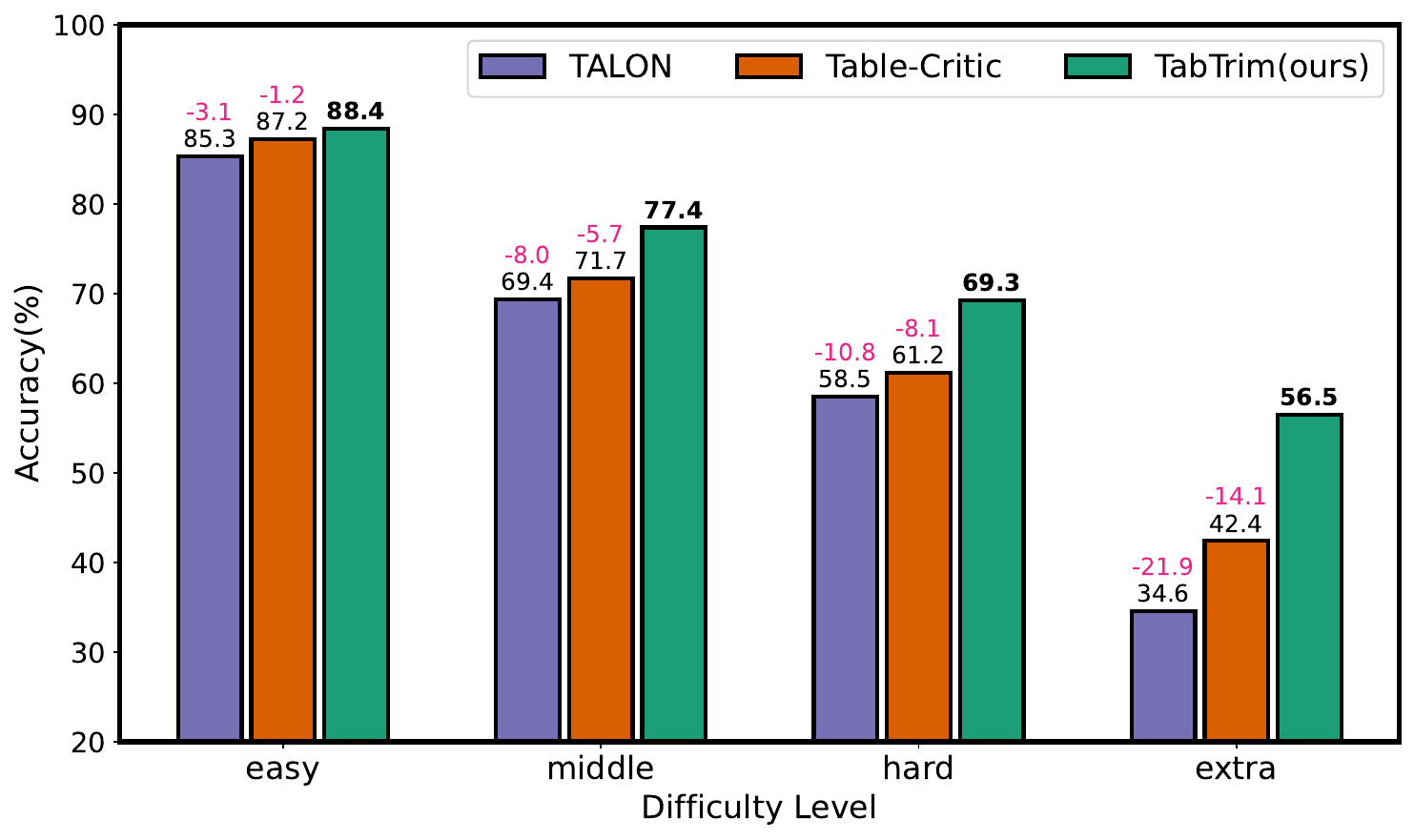}
    \caption{Accuracy comparison of TabTrim, Table-Critic, and TALON across different difficulty levels in the WikiTQ dataset.}
    \label{fig:hardness}
\end{figure}

\subsection{Fine-grained Analysis on Hardness}

To probe how TabTrim behaves on complex questions, we follow prior work~\citep{ye2025tableqa} and stratify WikiTQ questions into difficulty levels based on the empirical success rate of GPT-4o. Concretely, for each question we sample 100 independent answers from GPT-4o and count how many are correct, then assign a hardness label: Easy (90--100 correct), Medium (60--89), Hard (10--59), or Extra Hard (0--9). Fig.\ref{fig:hardness} compares TabTrim-8B against Table-Critic and TALON under these hardness buckets. TabTrim-8B consistently achieves higher accuracy across all levels, with the most pronounced improvements in the Extra Hard regime where baseline performance drops sharply. This trend suggests that TabTrim is especially beneficial for challenging questions, as it more reliably preserves answer-critical data needed for downstream multi-step reasoning.

\begin{table}[t]
  \centering
  \small
  \setlength{\tabcolsep}{6pt}
  \resizebox{\linewidth}{!}{
  \begin{tabular}{l|cc|cc}
    \toprule
    Method & WikiTQ & $\triangledown$ & TableBench & $\triangledown$ \\
    \midrule
    TabTrim & \textbf{79.4} & -- &  \textbf{61.2} & --   \\
    \quad w/o DPO & 78.1  & \textcolor{purple}{-1.3} & 58.6 & \textcolor{purple}{-2.6} \\
    \quad w/o Correction Samples & 74.8  & \textcolor{purple}{-4.6} & 55.4 & \textcolor{purple}{-5.8} \\
    \quad w/o Training & 54.7  & \textcolor{purple}{-24.7} & 49.6 & \textcolor{purple}{-11.6} \\
    \bottomrule
  \end{tabular}
  }
  \caption{Ablation on the training pipeline of the sub-table pruner. $\triangledown$ denotes absolute accuracy degradation.}
  \label{tab:abl_train}
\end{table}

\begin{table}[t]
  \centering
  \small
  \setlength{\tabcolsep}{6pt}
  \resizebox{\linewidth}{!}{
  \begin{tabular}{l|cc|cc}
    \toprule
    Method & WikiTQ & $\triangledown$ & TableBench & $\triangledown$ \\
    \midrule
    Loss-aware score & \textbf{79.4} & -- &  \textbf{61.2} & --  \\
    Balanced score & 77.8  & \textcolor{purple}{-1.6} & 58.3 & \textcolor{purple}{-2.9} \\
    \bottomrule
  \end{tabular}
  }
  \caption{Effect of loss-aware verification. $\triangledown$ denotes absolute accuracy degradation.}
  \label{tab:abl_veri}
\end{table}

\begin{table}[t]
  \centering
  \small
  \setlength{\tabcolsep}{6pt}
  \resizebox{\linewidth}{!}{
  \begin{tabular}{l|cc|cc}
    \toprule
    Method & WikiTQ & $\triangledown$ & TableBench & $\triangledown$ \\
    \midrule
    Rank by Verifier     & \textbf{79.4} & --                   & \textbf{61.2} & --                   \\
    Rank by Likelihood   & 74.2          & \textcolor{purple}{-5.2} & 56.7          & \textcolor{purple}{-4.5} \\
    Sequential Revisions      & 72.9           & \textcolor{purple}{-6.5}  & 55.1           & \textcolor{purple}{-6.1}  \\
    \bottomrule
  \end{tabular}
  }
  \caption{Ablation on search and ranking signal in score-guided inference. $\triangledown$ denotes absolute accuracy degradation.}
  \label{tab:abl_search}
\end{table}

\subsection{Ablation Study}

We conduct ablation studies to quantify the contribution of each component in TabTrim. Unless otherwise stated, all ablations are conducted on TabTrim-8B under the same inference configuration as the main results, and we report accuracy on WikiTQ and TableBench (total EM).

\paragraph{Effect of Pruner Training.}
We first ablate the training recipe of the pruner. Starting from the full model, we progressively remove (1) DPO, (2) correction samples constructed from modified off-trajectory sub-tables, and (3) all training (i.e., using the base model without fine-tuning). Tab.\ref{tab:abl_train} shows a consistent degradation as supervision is removed: dropping DPO reduces performance on both datasets, and removing correction samples further harms accuracy, indicating that robustness to off-trajectory sub-tables is crucial for reliable multi-step pruning. When training is removed entirely, performance collapses (-24.7\% on WikiTQ and -11.6\% on TableBench), approaching the direct QA baseline. This confirms that process-supervised trajectories are essential for learning effective pruning behaviors.

\paragraph{Effect of Loss-aware Verification.}
Next, we examine whether the verifier's loss-aware scoring is necessary. We compare our default setting ($\alpha=1.5$) with a balanced variant ($\alpha=1$) that reduces the emphasis on retaining answer-critical data. As shown in Tab.\ref{tab:abl_veri}, using the balanced score consistently degrades performance (-1.6\% on WikiTQ and -2.9\% on TableBench), indicating that a loss-aware quality score provides a stronger keep/discard criterion for multi-step pruning.

\paragraph{Effect of Parallel Trajectory Search.}
Finally, we analyze the role of Parallel Trajectory Search and its ranking signal. Tab.\ref{tab:abl_search} compares (1) the full TabTrim that performs beam search and ranks candidates by the verifier score, (2) a variant that keeps the same beam search but ranks by the pruner likelihood, and (3) the sequential revisions that disables search (i.e., $k=b=1$) with a matched call budget by increasing the maximum depth accordingly. Replacing verifier-based ranking with likelihood ranking causes substantial drops on both datasets (-5.2\% on WikiTQ and -4.5\% on TableBench), suggesting that likelihood is not a reliable proxy for sub-table quality. Moreover, disabling search also degrades performance (-6.5\% on WikiTQ and -6.1\% on TableBench), confirming that expanding beyond sequential revisions is important for recovering from early pruning errors.

\subsection{Analysis of Scaling}

\begin{figure}[t]
  \centering
  \includegraphics[width=\linewidth]{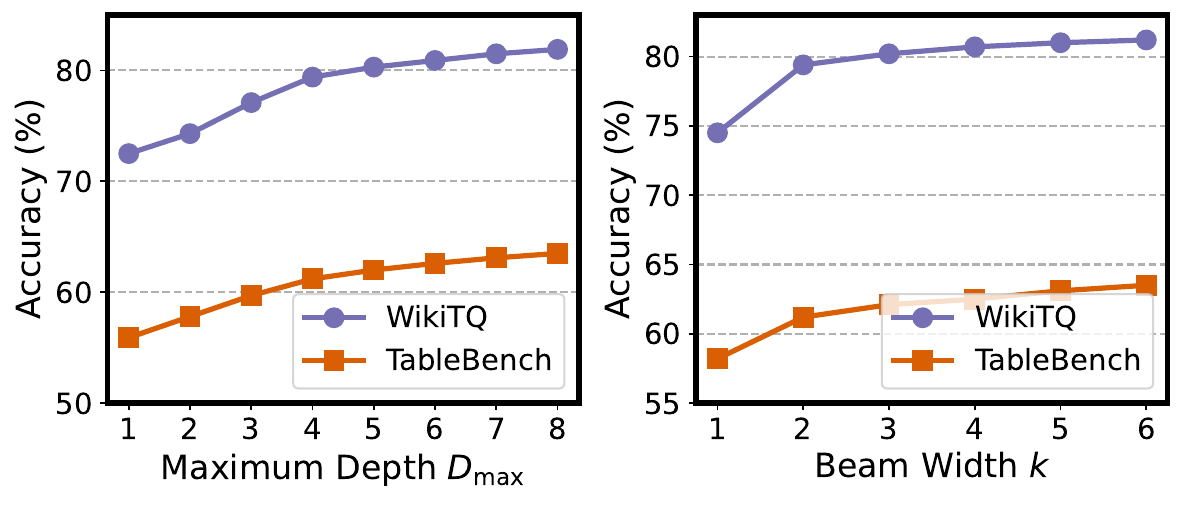}
  \caption{Scaling behavior of TabTrim-8B with increased inference-time search budget on WikiTQ and TableBench. Left: varying maximum depth $D_{\max}$ with $k=b=2$. Right: varying beam width $k$ with $D_{\max}=4$ and $b=2$.}
  \label{scaling:a}
\end{figure}

We analyze how TabTrim scales with increased inference-time compute. Fig.\ref{scaling:a} reports the performance of TabTrim-8B on WikiTQ and TableBench under varying search configurations. When we fix the beam width and branching factor ($k=b=2$) and gradually increase the maximum depth $D_{\max}$, accuracy improves monotonically from the shallow setting ($D_{\max}=1$) to deeper searches on both datasets. Similarly, when we fix $D_{\max}=4$ and branching factor $b=2$ and increase the beam width (from $k=1$ to larger values), performance also improves, indicating that TabTrim can reliably convert additional search budget into accuracy gains rather than becoming unstable.

\begin{figure}[t]
    \centering
    \includegraphics[width=\linewidth]{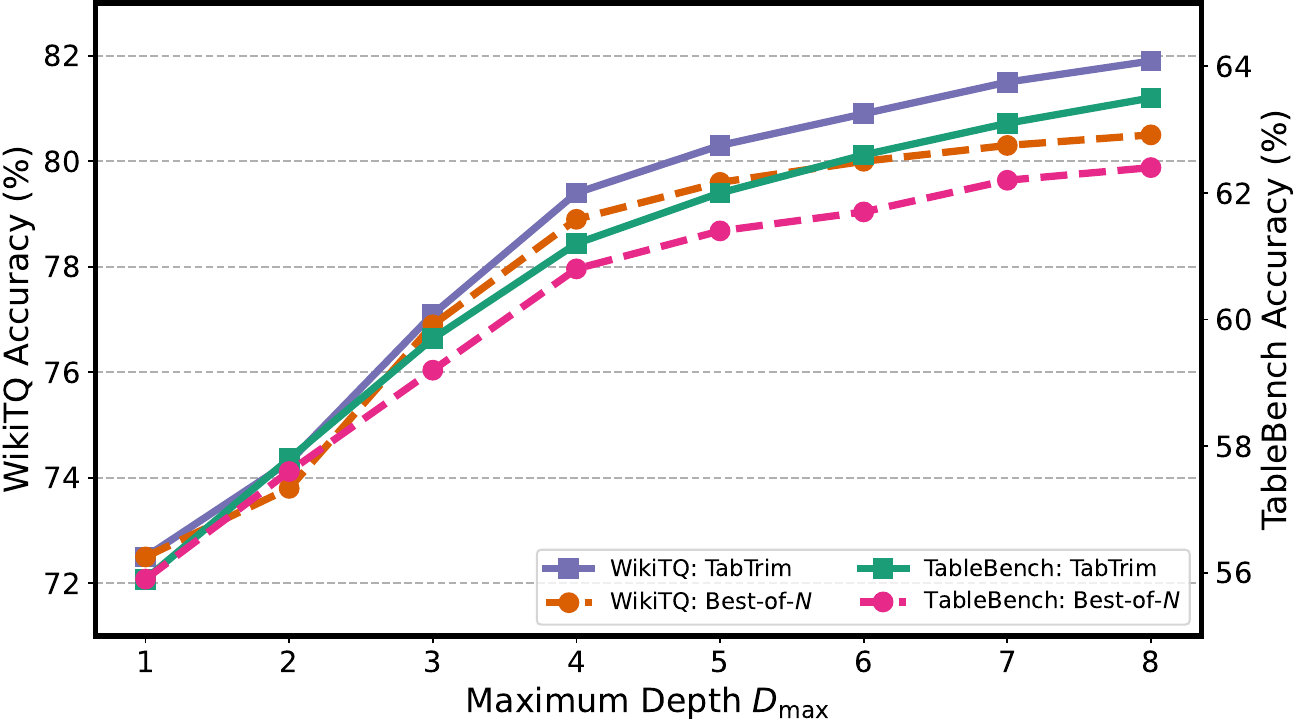}
    \caption{TabTrim vs.\ Best-of-$N$ sampling under a matched compute budget. We fix $k=b=2$ for TabTrim and set Best-of-$N$ to $N=k \cdot b=4$ and vary the maximum depth $D_{\max}$. We report accuracy on WikiTQ and TableBench.}
    \label{scaling:b}
\end{figure}

We further compare Parallel Trajectory Search with naive sampling under a matched compute budget. Specifically, we construct a Best-of-$N$ baseline that uses the same pruner and verifier, but replaces beam search with $N$ independent $D_{\max}$-step pruning rollouts and returns the rollout with the highest verifier score (we set $N=k \cdot b$ to match the per-depth expansion budget). As shown in Fig.\ref{scaling:b}, TabTrim consistently outperforms Best-of-$N$ on both WikiTQ and TableBench across all depths. This indicates that maintaining and selecting sub-tables step-by-step during the search is more effective than allocating the same calls to independent rollouts and only selecting at the end. Overall, these results demonstrate that TabTrim scales positively with inference-time compute and uses the search budget more efficiently than naive sampling.

\subsection{Analysis of Plug-and-Play Gains}

As TabTrim is an open-source, plug-and-play pruning module, we evaluate its plug-and-play gains on downstream open-source reasoners. Concretely, we run Qwen3-8B and Table-R1-7B \citep{yang-etal-2025-table} on (1) raw tables and (2) TabTrim(8B)-selected sub-tables, and report accuracy on WikiTQ and TableBench. Tab.\ref{tab:compatibility} shows that TabTrim consistently boosts both reasoners across datasets, indicating that TabTrim can serve as an effective front-end that reduces distracting noise while preserving answer-critical data for downstream reasoning.

\begin{table}[t]
  \centering
  \small
  \setlength{\tabcolsep}{6pt}
  \resizebox{\linewidth}{!}{
  \begin{tabular}{l|cc|cc}
    \toprule
    Method & WikiTQ & $\triangle$ & TableBench & $\triangle$ \\
    \midrule
    Qwen3                 & 52.2 & --  & 48.4 & -- \\
    \quad w/ TabTrim & 78.1 & \textcolor{teal}{+25.9} & 59.2 & \textcolor{teal}{+10.8} \\
    \midrule
    Table-R1                 & 77.5 & --  & 34.3 & -- \\
    \quad w/ TabTrim & 84.9 & \textcolor{teal}{+7.4} & 43.1 & \textcolor{teal}{+8.8} \\
    \bottomrule
  \end{tabular}
  }
  \caption{Plug-and-play gains from using TabTrim-selected sub-tables for open-source downstream reasoners. $\triangle$ denotes absolute accuracy gain.}
  \label{tab:compatibility}
\end{table}

\subsection{Analysis of Computational Cost}
\label{sec:cost}

We evaluate the computational cost of TabTrim using token counts (in millions) for both input and output, as shown in Tab.\ref{tab:token_cost}. Compared to the Table-Critic, TabTrim achieves significantly higher efficiency, consuming substantially fewer input tokens. Although TabTrim generates more output tokens due to its candidate generation mechanism, its total token consumption remains markedly lower, ranging from $0.56\times$ to $0.66\times$ of the TableCritic's total. These results demonstrate that TabTrim not only provides superior pruning performance but also maintains a much higher aggregate compute efficiency.

\section{Related Work}

\paragraph{Table Reasoning.}
Early work on table reasoning develops task-specific models with table-aware pretraining or fine-tuning objectives \citep{herzig-etal-2020-tapas, yin-etal-2020-tabert, 10.1145/3542700.3542709, liu2022tapex, gu-etal-2022-pasta}. With the rise of general-purpose LLMs, recent methods often serialize tables into text and perform open-text reasoning \citep{zhang2025survey} and introduce decomposition strategies to handle complex queries \citep{zhao-etal-2024-tapera, wu-feng-2024-protrix}. However, purely textual reasoning can be brittle as irrelevant noisy content can distract the model and undermine reasoning.

\paragraph{Table Pruning.}
To filter out noise before reasoning, recent work increasingly focuses on table pruning. Program-based methods rely on executable SQL or Python to filter rows and columns \citep{Binder, 10.14778/3659437.3659452, nahid-rafiei-2024-tabsqlify, jin-etal-2025-talon}, and LLM-based methods perform pruning through iterative reasoning \citep{ye2023large, wangchain, yu-etal-2025-table}. However, these methods are typically error-sensitive, where an early-stage pruning error can propagate and lead to reasoning failure.

\paragraph{Critique Mechanism.}
To alleviate error propagation, recent work explores critique mechanisms \citep{madaan2023self, yang-etal-2025-confidence} to evaluate intermediate steps and revise subsequent decisions. In table pruning, program-based methods often rely on execution feedback to trigger repair or regeneration~\citep{jin-etal-2025-talon}, while LLM-based methods commonly use LLM-as-a-Judge to assess intermediate outputs \citep{yu-etal-2025-table}. However, these critique signals are not reliable with true pruning errors, and these methods typically applied in a sequential revision process. As a result, erroneous pruning decisions may go undetected, and once answer-critical data is discarded, later revisions are often unable to recover it. To address these limitations, TabTrim derives step-wise supervision from sub-table trajectories to train a pruner and a verifier, and performs inference via parallel search to maintain multiple competing pruning trajectories and discard low-quality branches.

\begin{table}[t]
\centering
\small
\setlength{\tabcolsep}{5pt}
\resizebox{\linewidth}{!}{
\begin{tabular}{l|cc|cc|ccc}
\toprule
\multirow{2}{*}{Dataset} & \multicolumn{2}{c|}{Table-Critic} & \multicolumn{2}{c|}{TabTrim} & \multirow{2}{*}{Ratio} \\
 & Input (M) & Output (M) & Input (M) & Output (M) &  \\
\midrule
WikiTQ  & 135.5 & 3.8 & 63.8 & 14.5 & 0.56$\times$ \\
TabFact & 62.1  & 2.4 & 32.9 & 9.7  & 0.66$\times$ \\
\bottomrule
\end{tabular}
}
\caption{Token usage over WikiTQ and TabFact.}
\label{tab:token_cost}
\end{table}

\section{Conclusion}

In this paper, we propose TabTrim, a framework that performs step-wise pruning with reliable critique through parallel search, reducing answer-critical data loss while extracting simplified sub-tables to enhance downstream reasoning. TabTrim derives step-wise supervision from sub-table trajectories to train a pruner that generates pruned sub-tables and a verifier that scores intermediate sub-tables. At inference, TabTrim performs parallel search to maintain multiple competing pruning trajectories and discard low-quality branches. Experiments on WikiTQ, TabFact, and TableBench demonstrate that TabTrim consistently outperforms strong baselines, effectively improving table pruning and enhancing downstream table reasoning.

\section*{Limitations}

Our experimental scope is currently bounded by computational resource constraints, confining our evaluation to the 4B and 8B parameter regimes. While TabTrim demonstrates superior performance at this scale, the substantial computational expenditure required for fine-tuning massive architectures precludes the extension of our analysis to the larger range. Consequently, the scaling laws and performance ceiling of TabTrim with stronger base models remain to be fully characterized, an investigation we leave for future work.

\section*{Ethics Statement}

Our work aims to enhance the reliability of multi-step table pruning in TableQA. However, like any system based on LLMs, it still entails the risk of generating factually incorrect answers or incomplete sub-tables. We strongly advise users to exercise caution and verify critical outputs when deploying TabTrim in real-world scenarios. Furthermore, our research builds upon open-source advancements, specifically utilizing models such as Qwen3 and frameworks including PyTorch and Hugging Face. We strictly adhere to the respective licenses and usage policies of these resources, acknowledging their pivotal contribution to the community.

\section*{Acknowledgement}

This work was financially supported by National Key Research and Development Program of China, No.2024YDLN0004 and the National Natural Science Foundation of China (U23A20275).


\bibliography{custom}

@article{zhang2025survey,
  title={A survey of table reasoning with large language models},
  author={Zhang, Xuanliang and Wang, Dingzirui and Dou, Longxu and Zhu, Qingfu and Che, Wanxiang},
  journal={Frontiers of Computer Science},
  volume={19},
  number={9},
  pages={199348},
  year={2025},
  publisher={Springer}
}

@article{chen2024tablerag,
  title={Tablerag: Million-token table understanding with language models},
  author={Chen, Si-An and Miculicich, Lesly and Eisenschlos, Julian and Wang, Zifeng and Wang, Zilong and Chen, Yanfei and Fujii, Yasuhisa and Lin, Hsuan-Tien and Lee, Chen-Yu and Pfister, Tomas},
  journal={Advances in Neural Information Processing Systems},
  volume={37},
  pages={74899--74921},
  year={2024}
}

@article{Binder,
  title={Binding Language Models in Symbolic Languages},
  author={Zhoujun Cheng and Tianbao Xie and Peng Shi and Chengzu Li and Rahul Nadkarni and Yushi Hu and Caiming Xiong and Dragomir Radev and Mari Ostendorf and Luke Zettlemoyer and Noah A. Smith and Tao Yu},
  journal={ICLR},
  year={2023}
}

@inproceedings{ye2023large,
  title={Large language models are versatile decomposers: Decomposing evidence and questions for table-based reasoning},
  author={Ye, Yunhu and Hui, Binyuan and Yang, Min and Li, Binhua and Huang, Fei and Li, Yongbin},
  booktitle={Proceedings of the 46th international ACM SIGIR conference on research and development in information retrieval},
  pages={174--184},
  year={2023}
}

@inproceedings{wangchain,
  title={Chain-of-Table: Evolving Tables in the Reasoning Chain for Table Understanding},
  author={Wang, Zilong and Zhang, Hao and Li, Chun-Liang and Eisenschlos, Julian Martin and Perot, Vincent and Wang, Zifeng and Miculicich, Lesly and Fujii, Yasuhisa and Shang, Jingbo and Lee, Chen-Yu and Pfister, Tomas},
  booktitle={The Twelfth International Conference on Learning Representations},
  year={2024}
}

@article{10.14778/3659437.3659452,
author = {Zhang, Yunjia and Henkel, Jordan and Floratou, Avrilia and Cahoon, Joyce and Deep, Shaleen and Patel, Jignesh M.},
title = {ReAcTable: Enhancing ReAct for Table Question Answering},
year = {2024},
issue_date = {April 2024},
publisher = {VLDB Endowment},
volume = {17},
number = {8},
issn = {2150-8097},
journal = {Proc. VLDB Endow.},
month = apr,
pages = {1981–1994},
numpages = {14}
}

@inproceedings{nahid-rafiei-2024-tabsqlify,
    title = "{T}ab{SQL}ify: Enhancing Reasoning Capabilities of {LLM}s Through Table Decomposition",
    author = "Nahid, Md Mahadi Hasan  and
      Rafiei, Davood",
    editor = "Duh, Kevin  and
      Gomez, Helena  and
      Bethard, Steven",
    booktitle = "Proceedings of the 2024 Conference of the North American Chapter of the Association for Computational Linguistics: Human Language Technologies (Volume 1: Long Papers)",
    month = jun,
    year = "2024",
    address = "Mexico City, Mexico",
    publisher = "Association for Computational Linguistics",
    pages = "5725--5737",
}

@inproceedings{huanglarge,
  title={Large Language Models Cannot Self-Correct Reasoning Yet},
  author={Huang, Jie and Chen, Xinyun and Mishra, Swaroop and Zheng, Huaixiu Steven and Yu, Adams Wei and Song, Xinying and Zhou, Denny},
  year={2024},
  booktitle={The Twelfth International Conference on Learning Representations}
}

@inproceedings{yu-etal-2025-table,
    title = "Table-Critic: A Multi-Agent Framework for Collaborative Criticism and Refinement in Table Reasoning",
    author = "Yu, Peiying  and
      Chen, Guoxin  and
      Wang, Jingjing",
    editor = "Che, Wanxiang  and
      Nabende, Joyce  and
      Shutova, Ekaterina  and
      Pilehvar, Mohammad Taher",
    booktitle = "Proceedings of the 63rd Annual Meeting of the Association for Computational Linguistics (Volume 1: Long Papers)",
    month = jul,
    year = "2025",
    address = "Vienna, Austria",
    publisher = "Association for Computational Linguistics",
    pages = "17432--17451",
    ISBN = "979-8-89176-251-0",
}

@article{rafailov2023direct,
  title={Direct preference optimization: Your language model is secretly a reward model},
  author={Rafailov, Rafael and Sharma, Archit and Mitchell, Eric and Manning, Christopher D and Ermon, Stefano and Finn, Chelsea},
  journal={Advances in neural information processing systems},
  volume={36},
  pages={53728--53741},
  year={2023}
}

@inproceedings{wu-etal-2023-tacr,
    title = "{TACR}: A Table Alignment-based Cell Selection Method for {H}ybrid{QA}",
    author = {Wu, Jian  and
      Xu, Yicheng  and
      Gao, Yan  and
      Lou, Jian-Guang  and
      Karlsson, B{\"o}rje F.  and
      Okumura, Manabu},
    editor = "Rogers, Anna  and
      Boyd-Graber, Jordan  and
      Okazaki, Naoaki",
    booktitle = "Findings of the Association for Computational Linguistics: ACL 2023",
    month = jul,
    year = "2023",
    address = "Toronto, Canada",
    publisher = "Association for Computational Linguistics",
    pages = "6535--6549",
}

@inproceedings{lin-etal-2023-inner,
    title = "An Inner Table Retriever for Robust Table Question Answering",
    author = "Lin, Weizhe  and
      Blloshmi, Rexhina  and
      Byrne, Bill  and
      de Gispert, Adria  and
      Iglesias, Gonzalo",
    editor = "Rogers, Anna  and
      Boyd-Graber, Jordan  and
      Okazaki, Naoaki",
    booktitle = "Proceedings of the 61st Annual Meeting of the Association for Computational Linguistics (Volume 1: Long Papers)",
    month = jul,
    year = "2023",
    address = "Toronto, Canada",
    publisher = "Association for Computational Linguistics",
    pages = "9909--9926",
}

@article{zhong2017seq2sql,
  title={Seq2sql: Generating structured queries from natural language using reinforcement learning},
  author={Zhong, Victor and Xiong, Caiming and Socher, Richard},
  journal={arXiv preprint arXiv:1709.00103},
  year={2017}
}

@article{yang2025qwen3,
  title={Qwen3 technical report},
  author={Yang, An and Li, Anfeng and Yang, Baosong and Zhang, Beichen and Hui, Binyuan and Zheng, Bo and Yu, Bowen and Gao, Chang and Huang, Chengen and Lv, Chenxu and others},
  journal={arXiv preprint arXiv:2505.09388},
  year={2025}
}

@inproceedings{wu2025tablebench,
  title={Tablebench: A comprehensive and complex benchmark for table question answering},
  author={Wu, Xianjie and Yang, Jian and Chai, Linzheng and Zhang, Ge and Liu, Jiaheng and Du, Xeron and Liang, Di and Shu, Daixin and Cheng, Xianfu and Sun, Tianzhen and others},
  booktitle={Proceedings of the AAAI Conference on Artificial Intelligence},
  volume={39},
  number={24},
  pages={25497--25506},
  year={2025}
}

@inproceedings{pasupat-liang-2015-compositional,
    title = "Compositional Semantic Parsing on Semi-Structured Tables",
    author = "Pasupat, Panupong  and
      Liang, Percy",
    editor = "Zong, Chengqing  and
      Strube, Michael",
    booktitle = "Proceedings of the 53rd Annual Meeting of the Association for Computational Linguistics and the 7th International Joint Conference on Natural Language Processing (Volume 1: Long Papers)",
    month = jul,
    year = "2015",
    address = "Beijing, China",
    publisher = "Association for Computational Linguistics",
    pages = "1470--1480"
}

@inproceedings{2019TabFactA,
  title={TabFact : A Large-scale Dataset for Table-based Fact Verification},
  author={Chen, Wenhu and Wang, Hongmin and Chen, Jianshu and Zhang, Yunkai and Wang, Hong and Li, Shiyang and Zhou, Xiyou and Wang, William Yang},
  booktitle = {International Conference on Learning Representations (ICLR)},
  address = {Addis Ababa, Ethiopia},
  month = {April},
  year = {2020}
}

@article{hurst2024gpt,
  title={Gpt-4o system card},
  author={Hurst, Aaron and Lerer, Adam and Goucher, Adam P and Perelman, Adam and Ramesh, Aditya and Clark, Aidan and Ostrow, AJ and Welihinda, Akila and Hayes, Alan and Radford, Alec and others},
  journal={arXiv preprint arXiv:2410.21276},
  year={2024}
}

@article{ye2025tableqa,
  title={When TableQA Meets Noise: A Dual Denoising Framework for Complex Questions and Large-scale Tables},
  author={Ye, Shenghao and Guo, Yu and Jin, Dong and Shen, Yikai and Hou, Yunpeng and Chen, Shuangwu and Yang, Jian and Jiang, Xiaofeng},
  journal={arXiv preprint arXiv:2509.17680},
  year={2025}
}

@inproceedings{shi-etal-2020-potential,
    title = "On the Potential of Lexico-logical Alignments for Semantic Parsing to {SQL} Queries",
    author = "Shi, Tianze  and
      Zhao, Chen  and
      Boyd-Graber, Jordan  and
      Daum{\'e} III, Hal  and
      Lee, Lillian",
    editor = "Cohn, Trevor  and
      He, Yulan  and
      Liu, Yang",
    booktitle = "Findings of the Association for Computational Linguistics: EMNLP 2020",
    month = nov,
    year = "2020",
    address = "Online",
    publisher = "Association for Computational Linguistics",
    pages = "1849--1864",
}

@inproceedings{herzig-etal-2020-tapas,
    title = "{T}a{P}as: Weakly Supervised Table Parsing via Pre-training",
    author = {Herzig, Jonathan  and
      Nowak, Pawel Krzysztof  and
      M{\"u}ller, Thomas  and
      Piccinno, Francesco  and
      Eisenschlos, Julian},
    editor = "Jurafsky, Dan  and
      Chai, Joyce  and
      Schluter, Natalie  and
      Tetreault, Joel",
    booktitle = "Proceedings of the 58th Annual Meeting of the Association for Computational Linguistics",
    month = jul,
    year = "2020",
    address = "Online",
    publisher = "Association for Computational Linguistics",
    pages = "4320--4333",
}

@inproceedings{yin-etal-2020-tabert,
    title = "{T}a{BERT}: Pretraining for Joint Understanding of Textual and Tabular Data",
    author = "Yin, Pengcheng  and
      Neubig, Graham  and
      Yih, Wen-tau  and
      Riedel, Sebastian",
    editor = "Jurafsky, Dan  and
      Chai, Joyce  and
      Schluter, Natalie  and
      Tetreault, Joel",
    booktitle = "Proceedings of the 58th Annual Meeting of the Association for Computational Linguistics",
    month = jul,
    year = "2020",
    address = "Online",
    publisher = "Association for Computational Linguistics",
    pages = "8413--8426",
}

@article{10.1145/3542700.3542709,
author = {Deng, Xiang and Sun, Huan and Lees, Alyssa and Wu, You and Yu, Cong},
title = {TURL: Table Understanding through Representation Learning},
year = {2022},
issue_date = {March 2022},
publisher = {Association for Computing Machinery},
address = {New York, NY, USA},
volume = {51},
number = {1},
issn = {0163-5808},
journal = {SIGMOD Rec.},
month = jun,
pages = {33–40},
numpages = {8}
}

@inproceedings{
    liu2022tapex,
    title={{TAPEX}: Table Pre-training via Learning a Neural {SQL} Executor},
    author={Qian Liu and Bei Chen and Jiaqi Guo and Morteza Ziyadi and Zeqi Lin and Weizhu Chen and Jian-Guang Lou},
    booktitle={International Conference on Learning Representations},
    year={2022},
}

@inproceedings{gu-etal-2022-pasta,
    title = "{PASTA}: Table-Operations Aware Fact Verification via Sentence-Table Cloze Pre-training",
    author = "Gu, Zihui  and
      Fan, Ju  and
      Tang, Nan  and
      Nakov, Preslav  and
      Zhao, Xiaoman  and
      Du, Xiaoyong",
    editor = "Goldberg, Yoav  and
      Kozareva, Zornitsa  and
      Zhang, Yue",
    booktitle = "Proceedings of the 2022 Conference on Empirical Methods in Natural Language Processing",
    month = dec,
    year = "2022",
    address = "Abu Dhabi, United Arab Emirates",
    publisher = "Association for Computational Linguistics",
    pages = "4971--4983",
}

@inproceedings{zhao-etal-2024-tapera,
    title = "{T}a{PERA}: Enhancing Faithfulness and Interpretability in Long-Form Table {QA} by Content Planning and Execution-based Reasoning",
    author = "Zhao, Yilun  and
      Chen, Lyuhao  and
      Cohan, Arman  and
      Zhao, Chen",
    editor = "Ku, Lun-Wei  and
      Martins, Andre  and
      Srikumar, Vivek",
    booktitle = "Proceedings of the 62nd Annual Meeting of the Association for Computational Linguistics (Volume 1: Long Papers)",
    month = aug,
    year = "2024",
    address = "Bangkok, Thailand",
    publisher = "Association for Computational Linguistics",
    pages = "12824--12840",
}

@inproceedings{wu-feng-2024-protrix,
    title = "{P}ro{T}rix: Building Models for Planning and Reasoning over Tables with Sentence Context",
    author = "Wu, Zirui  and
      Feng, Yansong",
    editor = "Al-Onaizan, Yaser  and
      Bansal, Mohit  and
      Chen, Yun-Nung",
    booktitle = "Findings of the Association for Computational Linguistics: EMNLP 2024",
    month = nov,
    year = "2024",
    address = "Miami, Florida, USA",
    publisher = "Association for Computational Linguistics",
    pages = "4378--4406",
}

@inproceedings{yang-etal-2025-table,
    title = "Table-R1: Inference-Time Scaling for Table Reasoning Tasks",
    author = "Yang, Zheyuan  and
      Chen, Lyuhao  and
      Cohan, Arman  and
      Zhao, Yilun",
    editor = "Christodoulopoulos, Christos  and
      Chakraborty, Tanmoy  and
      Rose, Carolyn  and
      Peng, Violet",
    booktitle = "Proceedings of the 2025 Conference on Empirical Methods in Natural Language Processing",
    month = nov,
    year = "2025",
    address = "Suzhou, China",
    publisher = "Association for Computational Linguistics",
    pages = "20616--20635",
    ISBN = "979-8-89176-332-6",
}

@article{madaan2023self,
  title={Self-refine: Iterative refinement with self-feedback},
  author={Madaan, Aman and Tandon, Niket and Gupta, Prakhar and Hallinan, Skyler and Gao, Luyu and Wiegreffe, Sarah and Alon, Uri and Dziri, Nouha and Prabhumoye, Shrimai and Yang, Yiming and others},
  journal={Advances in Neural Information Processing Systems},
  volume={36},
  pages={46534--46594},
  year={2023}
}

@inproceedings{yang-etal-2025-confidence,
    title = "Confidence v.s. Critique: A Decomposition of Self-Correction Capability for {LLM}s",
    author = "Yang, Zhe  and
      Zhang, Yichang  and
      Wang, Yudong  and
      Xu, Ziyao  and
      Lin, Junyang  and
      Sui, Zhifang",
    editor = "Che, Wanxiang  and
      Nabende, Joyce  and
      Shutova, Ekaterina  and
      Pilehvar, Mohammad Taher",
    booktitle = "Proceedings of the 63rd Annual Meeting of the Association for Computational Linguistics (Volume 1: Long Papers)",
    month = jul,
    year = "2025",
    address = "Vienna, Austria",
    publisher = "Association for Computational Linguistics",
    pages = "3998--4014",
    ISBN = "979-8-89176-251-0",
}

@inproceedings{jin-etal-2025-talon,
    title = "{TALON}: A Multi-Agent Framework for Long-Table Exploration and Question Answering",
    author = "Jin, Ruochun  and
      Wang, Xiyue  and
      Wang, Dong  and
      Zheng, Haoqi  and
      Qi, Yunpeng  and
      Yang, Silin  and
      Zhang, Meng",
    editor = "Christodoulopoulos, Christos  and
      Chakraborty, Tanmoy  and
      Rose, Carolyn  and
      Peng, Violet",
    booktitle = "Proceedings of the 2025 Conference on Empirical Methods in Natural Language Processing",
    month = nov,
    year = "2025",
    address = "Suzhou, China",
    publisher = "Association for Computational Linguistics",
    pages = "27385--27401",
    ISBN = "979-8-89176-332-6",
}

@article{liu2024synthvlm,
  title={Synthvlm: High-efficiency and high-quality synthetic data for vision language models},
  author={Liu, Zheng and Liang, Hao and Huang, Xijie and Xiong, Wentao and Yu, Qinhan and Sun, Linzhuang and Chen, Chong and He, Conghui and Cui, Bin and Zhang, Wentao},
  journal={arXiv preprint arXiv:2407.20756},
  volume={3},
  year={2024}
}

@article{liu2025uniform,
  title={From Uniform to Heterogeneous: Tailoring Policy Optimization to Every Token's Nature},
  author={Liu, Zheng and Liu, Mengjie and Wen, Siwei and Cai, Mengzhang and Cui, Bin and He, Conghui and Zhang, Wentao},
  journal={arXiv preprint arXiv:2509.16591},
  year={2025}
}

@article{liu2025fusion,
  title={FUSION: Fully integration of vision-language representations for deep cross-modal understanding},
  author={Liu, Zheng and Liu, Mengjie and Chen, Jingzhou and Xu, Jingwei and Cui, Bin and He, Conghui and Zhang, Wentao},
  journal={arXiv preprint arXiv:2504.09925},
  year={2025}
}

@article{lin2026scientific,
  title={Scientific Image Synthesis: Benchmarking, Methodologies, and Downstream Utility},
  author={Lin, Honglin and Qin, Chonghan and Liu, Zheng and Pei, Qizhi and Li, Yu and Zhong, Zhanping and Gao, Xin and Wang, Yanfeng and He, Conghui and Wu, Lijun},
  journal={arXiv preprint arXiv:2601.17027},
  year={2026}
}

@article{lin2026mmfinereason,
  title={MMFineReason: Closing the Multimodal Reasoning Gap via Open Data-Centric Methods},
  author={Lin, Honglin and Liu, Zheng and Zhu, Yun and Qin, Chonghan and Lin, Juekai and Shang, Xiaoran and He, Conghui and Zhang, Wentao and Wu, Lijun},
  journal={arXiv preprint arXiv:2601.21821},
  year={2026}
}

@article{liu2026chartverse,
  title={ChartVerse: Scaling Chart Reasoning via Reliable Programmatic Synthesis from Scratch},
  author={Liu, Zheng and Lin, Honglin and Qin, Chonghan and Wang, Xiaoyang and Gao, Xin and Li, Yu and Cai, Mengzhang and Zhu, Yun and Zhong, Zhanping and Pei, Qizhi and others},
  journal={arXiv preprint arXiv:2601.13606},
  year={2026}
}

@article{wu2026step,
  title={Step Potential Advantage Estimation: Harnessing Intermediate Confidence and Correctness for Efficient Mathematical Reasoning},
  author={Wu, Fei and Zhang, Zhenrong and Chang, Qikai and Zhang, Jianshu and Liu, Quan and Du, Jun},
  journal={arXiv preprint arXiv:2601.03823},
  year={2026}
}

@article{an2025amo,
  title={Amo-bench: Large language models still struggle in high school math competitions},
  author={An, Shengnan and Cai, Xunliang and Cao, Xuezhi and Li, Xiaoyu and Lin, Yehao and Liu, Junlin and Lv, Xinxuan and Ma, Dan and Wang, Xuanlin and Wang, Ziwen and others},
  journal={arXiv preprint arXiv:2510.26768},
  year={2025}
}

@article{Zhang2026ExpSeekSE,
  title={ExpSeek: Self-Triggered Experience Seeking for Web Agents},
  author={Wenyuan Zhang and Xinghua Zhang and Haiyang Yu and Shuaiyi Nie and Bingli Wu and Juwei Yue and Tingwen Liu and Yongbin Li},
  journal={arXiv preprint arXiv:2601.08605},
  year={2026},
}

@inproceedings{zhou2025dropping, title={Dropping Experts, Recombining Neurons: Retraining-Free Pruning for Sparse Mixture-of-Experts LLMs}, author={Zhou, Yixiao and Zhao, Ziyu and Cheng, Dongzhou and Wu, Zhiliang and Gui, Jie and Yang, Yi and Wu, Fei and Cheng, Yu and Fan, Hehe}, booktitle={Findings of the Association for Computational Linguistics: EMNLP 2025}, pages={15169--15186}, year={2025} }

@article{zhou2026look, title={Look Inward to Explore Outward: Learning Temperature Policy from LLM Internal States via Hierarchical RL}, author={Zhou, Yixiao and Li, Yang and Cheng, Dongzhou and Fan, Hehe and Cheng, Yu}, journal={arXiv preprint arXiv:2602.13035}, year={2026} }

@inproceedings{wu-etal-2025-ucs, title = "{UCS}-{SQL}: Uniting Content and Structure for Enhanced Semantic Bridging In Text-to-{SQL}", author = "Wu, Zhenhe and Li, Zhongqiu and Zhang, Jie and He, Zhongjiang and Yang, Jian and Zhao, Yu and Fang, Ruiyu and Wang, Bing and Xie, Hongyan and Song, Shuangyong and Li, Zhoujun", editor = "Che, Wanxiang and Nabende, Joyce and Shutova, Ekaterina and Pilehvar, Mohammad Taher", booktitle = "Findings of the Association for Computational Linguistics: ACL 2025", month = jul, year = "2025", address = "Vienna, Austria", publisher = "Association for Computational Linguistics", pages = "8156--8168", ISBN = "979-8-89176-256-5", }

@inproceedings{wu2025mr, title={MR-SQL: multi-level retrieval enhances inference for llm in text-to-sql}, author={Wu, Zhenhe and Li, Zhongqiu and Li, Mengxiang and Zhang, Jie and He, Zhongjiang and Yang, Jian and Zhao, Yu and Fang, Ruiyu and Li, Yongxiang and Li, Zhoujun and others}, booktitle={International Conference on Database Systems for Advanced Applications}, pages={403--413}, year={2025}, organization={Springer} }

@article{wu2025table,
  title={Table-r1: Region-based reinforcement learning for table understanding},
  author={Wu, Zhenhe and Yang, Jian and Liu, Jiaheng and Wu, Xianjie and Pan, Changzai and Zhang, Jie and Zhao, Yu and Song, Shuangyong and Li, Yongxiang and Li, Zhoujun},
  journal={arXiv preprint arXiv:2505.12415},
  year={2025}
}

\appendix

\section{Additional Implementation Details}
\label{sec:A} 

\subsection{SQL Decomposition}
\label{sec:decomposition}

Given a training triple $(Q, S_{\text{gold}}, T_{\text{raw}})$ from a Text-to-SQL dataset, we first parse the gold SQL query $S_{\text{gold}}$ into an abstract syntax tree using a SQL parser. We then linearize this tree into an ordered sequence of clause-level operations
\[
(c_1, c_2, \ldots, c_n),
\]
which follows the logical execution order of the query. In practice, we focus on clauses that directly affect the evidential content of the table. Concretely, we restrict intermediate operations to: (1) row selection based on predicates in \texttt{WHERE} (e.g., conjunctions or disjunctions over comparison operators), and (2) column pruning that determines the relevant subset of attributes in the base table (e.g., keeping columns that appear in \texttt{SELECT}, \texttt{GROUP BY}, \texttt{HAVING}, or aggregate expressions, while dropping attributes that never influence the answer), but implemented purely as column projection without changing row cardinality.

Starting from the full table $T_0 = T_{\text{raw}}$, each clause $c_t$ is applied as a transformation
\[
T_t^{+} = \text{Exec}(c_t, T_{t-1}^{+}),
\]
where $\text{Exec}(c_t,\cdot)$ programmatically applies the clause-level operation $c_t$ (e.g., filtering or projection) to the current sub-table to obtain the next sub-table. By construction, yielding an answer-critical data preserved pruning trajectory
\[
\tau^{+} = (T_0, T_1^{+}, \ldots, T_n^{+})
\]
that decomposes the pruning process into a sequence of sub-table. These trajectories serve as process supervision signals for training the pruner in Sec.\ref{sec:pruner}.

\subsection{SQL Modification}
\label{sec:modification}

To train the pruner and verifier to distinguish answer-critical sub-table from incorrect ones, we construct negative samples by modifying the clause sequence obtained in Appendix~\ref{sec:decomposition}. Given a training triple $(Q, SQL_{\text{gold}}, T_{\text{raw}})$ and its clause sequence $(c_1, \ldots, c_n)$, we generate modified clauses $\{c_t^{-}\}$ and obtain corresponding negative sub-tables $\{T_t^{-}\}$ by executing these clauses on the $\{T_{t-1}^{+}\}$.

\paragraph{Modification Space.}
We operate at the same granularity as our decomposition, i.e., on row-selection predicates and column-pruning decisions, so that all modifications remain executable and interpretable as pruning operations. Concretely, we use the following families of modifications:

\begin{itemize}
    \item \textbf{Predicate-level modifications.}
    For a \texttt{WHERE} clause predicate of the form $a \ \texttt{op} \ v$, we generate variants that (1) replace $v$ with another value from the same column domain (e.g., nearest neighbor in the sorted value set), or (2) swap $\texttt{op}$ with a nearby operator (e.g., $\texttt{<}$ vs.\ $\texttt{<=}$, $\texttt{>}$ vs.\ $\texttt{>=}$). For composite predicates connected by \texttt{AND}/\texttt{OR}, we either drop one conjunct (to admit more irrelevant rows) or add a spurious conjunct that filters out some answer-critical rows.
    \item \textbf{Column-pruning modifications.}
    For the column selection induced by $c_t$, we create perturbed versions by (1) dropping at least one attribute that participates in \texttt{SELECT}, \texttt{GROUP BY}, \texttt{HAVING}, or aggregate expressions, or (2) keeping additional irrelevant attributes. These are implemented as column projections on the table and do not change row cardinality.
\end{itemize}

All modified clauses are required to be syntactically valid and executable on $T_{\text{raw}}$. After execution, yielding modified sub-tables $T_t^{-}$ that are structurally comparable to the gold sub-table.

\paragraph{Negative Samples Construction.}
Starting from the gold trajectory $\tau^{+} = (T_0, T_1^{+}, \ldots, T_n^{+})$, we generate negative states in a step-wise manner. For each clause $c_t$, we sample one or more modifications from the above space to obtain $c_t^{-}$ and define
\[
T_t^{-} = \text{Exec}(c_t^{-}, T_{t-1}^+),
\]
where \text{Exec} denotes the executor described in Appendix~\ref{sec:decomposition}. We discard degenerate cases where $T_t^{-}$ is identical to $T_t^{+}$ or becomes trivially empty or full, and keep modifications that induce non-trivial but incorrect deviations from the on-trajectory sub-table.

\subsection{Cell-index Canonicalization}
\label{sec:canonicalization}

To make pruned sub-tables well-defined and comparable, we convert each raw table $T_{\text{raw}}$ into a set of indexed cells. All sub-table states $T_t$ in TabTrim are represented as subsets of this canonicalized cell set.

\paragraph{Canonical Cell Representation.}

We treat each table as a grid with row indices and column indices.
For each data row $r$ and column $c$ with value $v_{rc}$, we create a canonical cell
\[
x_{rc} = (\text{row}=r,\ \text{col}=c,\ \text{val}=v_{rc}),
\]
and define the canonicalized table as the cell set
\[
\mathcal{C}(T_{\text{raw}}) = \{x_{rc}\}.
\]
A sub-table state $T_t$ is then represented as a subset $\mathcal{C}(T_t) \subseteq \mathcal{C}(T_{\text{raw}})$ obtained by selecting a subset of rows and/or columns (equivalently, a subset of cells).

\section{Additional Experimental Details}
\label{sec:B}

\begin{table}[t]
  \centering
  \small
  \begin{tabular}{c|cc}
    \toprule
    $\lambda$ & WikiTQ & TableBench \\
    \midrule
    0.3  & 76.5 & 58.1 \\
    0.5  & 78.1 & 60.8 \\
    0.8  & 78.8 & 61.7 \\
    1.0  & 79.4 & 61.2 \\
    \bottomrule
  \end{tabular}
  \caption{Sensitivity analysis of the correction weight $\lambda$ in the pruner training objective.}
  \label{tab:lambda_sens}
\end{table}

\begin{table}[t]
  \centering
  \small
  \begin{tabular}{c|cc}
    \toprule
    $\alpha$ & WikiTQ & TableBench \\
    \midrule
    1.0  & 77.8 & 58.3 \\
    1.2  & 78.6 & 59.7 \\
    1.5  & 79.4 & 61.2 \\
    1.7  & 78.9 & 61.5 \\
    2.0  & 78.7 & 60.4 \\
    \bottomrule
  \end{tabular}
  \caption{Sensitivity analysis of the recall-bias parameter $\alpha$ in the loss-aware verifier score.}
  \label{tab:alpha_sens}
\end{table}

\subsection{Experiment Setup}
\label{sec:setup}

All experiments are conducted on 8 NVIDIA A100 40GB GPUs. We train all models for 2 epochs using the AdamW optimizer, employing a cosine learning rate schedule with a 3\% warmup ratio. For the Pruner (Qwen3-4B and Qwen3-8B), we perform full-parameter fine-tuning with a peak learning rate of $2\times 10^{-5}$ and set the correction weight to $\lambda=1$. In the subsequent DPO stage, we reduce the learning rate to $2\times 10^{-6}$ and set $\beta=0.2$, following standard DPO configurations. The Verifier (Qwen3-0.6B) is trained with a peak learning rate of $5\times 10^{-5}$. During inference, pruner generates outputs using top-p sampling of 0.95 and a temperature of 0.6, with a maximum generation length of 4,096 tokens. To ensure the stability of our results, we also repeated experiments with different random seeds. For TabTrim, we report the average accuracy over 3 runs. Across all datasets, the results were highly stable, with the standard deviation consistently within 1\% absolute accuracy.

\begin{table}[t]
\centering
\small
\begin{tabular}{lr}
\toprule
\textbf{Error Type} & \textbf{\%} \\
\midrule
\multicolumn{2}{l}{\textit{Sub-table errors}} \\
\quad Row mis-selection      & 5\% \\
\quad Column mis-selection   & 2\% \\
\midrule
\multicolumn{2}{l}{\textit{Reasoning errors}} \\
\quad Arithmetic error              & 42\% \\
\quad Aggregation/Counting error    & 38\% \\
\quad Misinterpretation             & 7\% \\
\quad Logical error                 & 2\% \\
\quad Others                        & 4\% \\
\bottomrule
\end{tabular}
\caption{Error taxonomy (percentage) on 100 failed examples of TabTrim-8B.}
\label{tab:error_taxonomy}
\end{table}

\begin{table}[t]
\centering
\small
\begin{tabular}{l|cc|c}
\toprule
\multirow{2}{*}{Dataset}  & \multicolumn{2}{c|}{\# Tokens per Table} &\multirow{2}{*}{Comp. (\%)} \\
\cmidrule(lr){2-3} 
& Entire Table & Pruned Table & \\
\midrule
TabFact & 353  & 215  & 39.1\%  \\
WikiTQ & 631  & 282   & 55.3\%  \\
\bottomrule
\end{tabular}
\caption{Token counts and compression rates before and after pruning across WikiTQ and TabFact datasets.}
\label{tab:compression}
\end{table}

\subsection{Dataset Details}
\label{sec:dataset}

\textbf{WikiTQ} introduces question answering over semi-structured HTML tables, aiming to test both compositional reasoning and domain generalization. It comprises 22,033 natural language questions paired with 2,108 Wikipedia tables, where the training and test tables are disjoint to ensure generalization to unseen schemas. The tables are semi-structured and heterogeneous, often containing multi-part cell values that require normalization into multiple semantic types such as numbers or dates. Questions range from simple lookups to highly compositional queries involving comparison, aggregation, arithmetic, and superlatives. Each table contains at least 8 rows and 5 columns, and the question collection was conducted with quality control through multiple annotators.

\textbf{TabFact} is a large-scale benchmark for table-based fact verification. Given a semi-structured Wikipedia table and a natural-language statement, the task is to predict whether the statement is ENTAILED or REFUTED by the table evidence. TabFact contains 118,275 human-annotated statements grounded in 16,573 Wikipedia tables. Each table with an average of 14 rows and 5-6 columns corresponds to 2--20 different statements, while each cell has an average of 2.1 words.

\textbf{TableBench} is a comprehensive benchmark specifically designed to evaluate the reasoning abilities of LLMs over tabular data. It consists of 3,681 unique tables drawn from diverse domains such as finance, sports, politics, and science, with each table containing on average 16.7 rows and 6.7 columns. The dataset emphasizes numerical reasoning, with over 65\% of table cells containing numerical values. TableBench questions are organized into four major categories: fact-checking, numerical reasoning, data analysis, further divided into 18 subcategories, yielding a total of 886 carefully annotated samples. Each question typically requires 6.3 reasoning steps, making the dataset significantly more complex than prior TableQA corpora.

To ensure a fair evaluation, we de-duplicate the training corpus and exclude any training tables that overlap with the tables in our test benchmarks.

\section{Additional Experiments}
\label{sec:C}

\subsection{Sensitivity to Correction Weight}
We analyze the correction weight $\lambda$ in the pruner training objective, which trades off progression learning on gold trajectories and recovery learning from off-trajectory states. We vary $\lambda$ while keeping the same training set, base model, optimization setup, and inference budget fixed, and report EM on WikiTQ and TableBench using TabTrim-8B in Tab.~\ref{tab:lambda_sens}. As $\lambda$ increases from 0.3 to 1.0, performance overall improves on both benchmarks, indicating that emphasizing recovery learning is important for mitigating off-trajectory errors. Moreover, the gains become marginal beyond $\lambda\approx 0.8$, suggesting that TabTrim is not overly sensitive to precise tuning within this moderate range.

\subsection{Sensitivity to Recall-bias Parameter}
We study the sensitivity of TabTrim to the recall-bias parameter $\alpha$ in the loss-aware verifier score. We keep the training data, base model, optimization setup, and inference budget fixed, and report EM on WikiTQ and TableBench using TabTrim-8B in Tab.~\ref{tab:alpha_sens}.
Overall, adopting a recall-biased setting ($\alpha>1$) consistently improves over the balanced score ($\alpha=1$), confirming the benefit of explicitly penalizing evidence loss. Performance peaks around $\alpha\in[1.5,1.7]$ and remains competitive for nearby values, suggesting that TabTrim only requires a moderate recall bias rather than precise tuning.

\subsection{Error Analysis}
We perform a fine-grained error analysis on 100 erroneous responses produced by TabTrim-8B. We manually categorize each failure into mutually exclusive error types based on whether the final pruned sub-table preserves answer-critical data, and if so, what type of reasoning mistake leads to the incorrect answer, the results are shown in Tab.\ref{tab:error_taxonomy}.

Overall, answer-critical selection errors are rare (7\%), indicating that TabTrim largely preserves answer-critical rows and columns in the pruned sub-table. The dominant failures arise from downstream reasoning, especially numerical computation (42\%) and aggregation/counting (38\%), suggesting that improving the answerer’s arithmetic and aggregation robustness is the primary direction for further gains.

\subsection{Deployment-oriented Cost Analysis}

While the raw token counts in Sec.\ref{sec:cost} demonstrate TabTrim's efficiency, its practical advantage is even more pronounced in self-hosted environments. TabTrim is designed on open-source models where prefix KV caching can amortize the cost of repeated prefills. Since calls to the pruner (and similarly the verifier) within the parallel search share a common prompt prefix (the query and raw table), this prefix is computed only once and reused across all branches. Besides, TabTrim's decoding overhead can be further reduced without changing the algorithm by adopting a more compact sub-table representation. Taken together, these considerations suggest that TabTrim can be even more favorable in practical self-hosted settings.

\subsection{Table Pruning Effectiveness}

To evaluate the effectiveness of TabTrim's table pruning capability, we randomly sample 200 correctly answered instances in WikiTQ and TabFact, and compute the average number of tokens before and after pruning. Table~\ref{tab:compression} reports the average token counts and corresponding compression rates.

\section{Training Data and Prompt}

\subsection{Example of training data for Pruner}

Tab.\ref{tab:pruner_data} presents the example of training data for the pruner.

\subsection{Example of training data for Verifier}

Tab.\ref{tab:verifier_data} presents the example of training data for the verifier.

\subsection{Prompt of Answerer}

Tab.\ref{tab:answerer} presents the prompt format used in answerer.

\section{Additional Related Work}

\paragraph{LLM Reasoning.} In recent years, the emergence of LLMs has established reasoning as a core capability for modern AI systems. Reasoning-based paradigms have been widely adopted across diverse domains, including vision-language modeling \citep{liu2024synthvlm,liu2025fusion,lin2026mmfinereason,lin2026scientific}, chart understanding \citep{liu2026chartverse}, mathematical problem solving \citep{wu2026step,an2025amo}, and web-based autonomous agents \citep{Zhang2026ExpSeekSE}. Concurrently, emerging evidence suggests that reasoning processes exhibit significant heterogeneity across different tokens, modules, modalities, and intermediate steps, calling for more adaptive reasoning, optimization, and inference strategies \citep{liu2025uniform,zhou2025dropping,zhou2026look}. Against this backdrop, structured reasoning tasks have attracted increasing attention, including table reasoning and closely related text-to-SQL settings, where models must jointly understand structured schemas, content, and compositional reasoning procedures over tabular evidence \citep{wu-etal-2025-ucs,wu2025mr,wu2025table}.

\begin{table*}[ht]
  \small
  \centering
  \begin{tabularx}{\textwidth}{>{\ttfamily\raggedright\arraybackslash}X}
    \hline
    \textbf{Example of training data for pruner}\tabularnewline
    \hline
    \#\#\# Question:\newline
    How many Engineering employees in Tokyo or Osaka were hired after 2020 and earn more than 130,000?\newline
    \#\#\# Raw Table:\newline
    col: Employee | Department | City | Hire Year | Salary | Level | Manager\newline
    row 1: Akira Sato | Engineering | Tokyo | 2021 | 120000 | L4 | K. Tanaka\newline
    row 2: Mei Chen | Engineering | Osaka | 2019 | 130000 | L4 | K. Tanaka\newline
    row 3: Haruto Ito | Engineering | Osaka | 2022 | 105000 | L3 | M. Suzuki\newline
    row 4: Yuna Park | Sales | Tokyo | 2021 | 90000 | L3 | R. Lee\newline
    row 5: Kenji Watanabe | Engineering | Nagoya | 2023 | 115000 | L4 | M. Suzuki\newline
    row 6: Sara Kim | Engineering | Tokyo | 2020 | 98000 | L3 | M. Suzuki\newline
    row 7: Rina Nakamura | Engineering | Tokyo | 2024 | 140000 | L5 | K. Tanaka\newline
    row 8: Daichi Mori | HR | Osaka | 2022 | 80000 | L2 | T. Yamada\newline
    ...\newline
    \#\#\# Current Sub-table:\newline
    (after Step 1: filter Department = Engineering; Step 2: filter City in {Tokyo, Osaka})\newline
    col: Employee | Department | City | Hire Year | Salary | Level | Manager\newline
    row 1: Akira Sato | Engineering | Tokyo | 2021 | 120000 | L4 | K. Tanaka\newline
    row 2: Mei Chen | Engineering | Osaka | 2019 | 130000 | L4 | K. Tanaka\newline
    row 3: Haruto Ito | Engineering | Osaka | 2022 | 105000 | L3 | M. Suzuki\newline
    row 6: Sara Kim | Engineering | Tokyo | 2020 | 98000 | L3 | M. Suzuki\newline
    row 7: Rina Nakamura | Engineering | Tokyo | 2024 | 140000 | L5 | K. Tanaka\newline
    ...\newline
    \#\#\# Next Sub-table:\newline
    (Step 3: filter Hire Year > 2020)\newline
    col: Employee | Department | City | Hire Year | Salary | Level | Manager\newline
    row 1: Akira Sato | Engineering | Tokyo | 2021 | 120000 | L4 | K. Tanaka\newline
    row 3: Haruto Ito | Engineering | Osaka | 2022 | 105000 | L3 | M. Suzuki\newline
    row 7: Rina Nakamura | Engineering | Tokyo | 2024 | 140000 | L5 | K. Tanaka\newline
    ...\tabularnewline
    \hline
  \end{tabularx}
  \caption{Example of training data for the pruner.}
  \label{tab:pruner_data}
\end{table*}

\begin{table*}[ht]
  \small
  \centering
  \begin{tabularx}{\textwidth}{>{\ttfamily\raggedright\arraybackslash}X}
    \hline
    \textbf{Example of training data for verifier}\tabularnewline
    \hline
    \#\#\# Question:\newline
    How many Engineering employees in Tokyo or Osaka were hired after 2020 and earn more than 130,000?\newline
    \#\#\# Raw Table:\newline
    col: Employee | Department | City | Hire Year | Salary | Level | Manager\newline
    row 1: Akira Sato | Engineering | Tokyo | 2021 | 120000 | L4 | K. Tanaka\newline
    row 2: Mei Chen | Engineering | Osaka | 2019 | 130000 | L4 | K. Tanaka\newline
    row 3: Haruto Ito | Engineering | Osaka | 2022 | 105000 | L3 | M. Suzuki\newline
    row 4: Yuna Park | Sales | Tokyo | 2021 | 90000 | L3 | R. Lee\newline
    row 5: Kenji Watanabe | Engineering | Nagoya | 2023 | 115000 | L4 | M. Suzuki\newline
    row 6: Sara Kim | Engineering | Tokyo | 2020 | 98000 | L3 | M. Suzuki\newline
    row 7: Rina Nakamura | Engineering | Tokyo | 2024 | 140000 | L5 | K. Tanaka\newline
    row 8: Daichi Mori | HR | Osaka | 2022 | 80000 | L2 | T. Yamada\newline
    ...\newline
    \#\#\# Sub-table:\newline
    (intermediate sub-table after Step 3)\newline
    col: Employee | Department | City | Hire Year | Salary | Level | Manager\newline
    row 1: Akira Sato | Engineering | Tokyo | 2021 | 120000 | L4 | K. Tanaka\newline
    row 3: Haruto Ito | Engineering | Osaka | 2022 | 105000 | L3 | M. Suzuki\newline
    row 7: Rina Nakamura | Engineering | Tokyo | 2024 | 140000 | L5 | K. Tanaka\newline
    ...\newline
    \#\#\# Score:\newline
    0.67\tabularnewline
    \hline
  \end{tabularx}
  \caption{Example of training data for the verifier.}
  \label{tab:verifier_data}
\end{table*}

\begin{table*}[ht]
  \small
  \centering
  \begin{tabularx}{\textwidth}{>{\ttfamily\raggedright\arraybackslash}X}
    \hline
    \textbf{Prompt of Answerer}\tabularnewline
    \hline
    \#\#\# Instruction:\newline
    You are a table question answering expert. Your task is to infer the answer to the question based on the provided table.\newline
    \#\#\# Question:\newline
    How many Engineering employees in Tokyo or Osaka were hired after 2020 and earn more than 130,000?\newline
    \#\#\# Table:\newline
    (final sub-table after Step 4: filter Salary > 130000; Step 5: keep only key columns)\newline
    col: Employee | City | Hire Year | Salary\newline
    row 7: Rina Nakamura | Tokyo | 2024 | 140000\newline
    \#\#\# Answer:\newline
    1\tabularnewline
    \hline
  \end{tabularx}
  \caption{Prompt format for the answerer.}
  \label{tab:answerer}
\end{table*}

\end{document}